\documentclass[10pt,notitlepage]{article}
\usepackage{epsfig}
\usepackage{amssymb}

\newcommand{\be}{\begin{equation}}
\newcommand{\ee}{\end{equation}}
\newcommand{\bea}{\begin{eqnarray}}
\newcommand{\eea}{\end{eqnarray}}

\usepackage{bm}
\usepackage{color}
\usepackage{amsmath}
\usepackage{graphicx}
\usepackage{amssymb}
\usepackage{cite}

\begin{document}

\title{On a novel training algorithm for sequence-to-sequence predictive recurrent networks}
\author{Boris  Rubinstein,
\\Stowers Institute for Medical Research
\\1000 50$^{}\mbox{th}$ St., Kansas City, MO 64110, U.S.A.}
\date{\today}
\maketitle

\begin{abstract}
Neural networks mapping sequences to sequences (seq2seq) lead to significant
progress in machine translation and speech recognition. Their traditional architecture
includes two recurrent networks (RNs) followed by a linear predictor.
In this manuscript we perform analysis of a corresponding algorithm and show that the parameters 
of the RNs of the well trained predictive network are not independent of each other. 
Their dependence can be used to significantly improve the network effectiveness.
The traditional seq2seq algorithms require short term memory
of a size proportional to the predicted sequence length. This requirement
is quite difficult to implement in a neuroscience context. 
We present a novel memoryless algorithm for seq2seq predictive networks
and compare it to the traditional one in the context of time series prediction.
We show that the new algorithm is more robust and makes predictions
with higher accuracy than the traditional one.
\end{abstract}

\section{Introduction}
The majority of predictive networks based of the recurrent networks (RNs) are designed
to use a fixed or variable length $m$ input sequence to
produce a single predicted element (all the input and an output element have the same structure). 
Such a system can be called $m$-to-$1$ predictive network.
It includes a chain of RNs (this chain can degenerate into a single RN) followed by a predictor
that converts a last inner state $\bm s_m$ of the last RN of the chain into the predicted element. 
In order to predict a sequence of elements one has to employ special algorithms that 
use the trained network recursively by appending already predicted terms to the 
input sequence. In an "expanding window" (EW) algorithm the length of the input sequence increases 
so that the network should be trained on the inputs of variable length.
To employ the input of fixed length one uses a "moving window" (MW) approach in which
after each prediction round the input sequence is modified by appending the predicted element
and dropping the first element of the current input. The recursive application of the $m$-to-$1$ network
for prediction of the element sequence requires an access to a short term memory 
to store the input sequence and this condition might be difficult to satisfy in neuroscience context. 
To resolve this problem the author recently suggested a
memoryless (ML) algorithm that was successfully applied for time series 
 prediction \cite{Rub2020a,Rub2020b}. 

The sequence prediction design can be considered from a different perspective
-- to construct a network that takes an input sequence and produces directly
an ordered {\it sequence} of $k$ predicted elements using sequence to sequence (seq2seq) algorithm.
This approach can also be called $m$-to-$k$ extension
of the  $m$-to-$1$ networks discussed above. Such seq2seq networks are considered to 
be an ideal tool for machine translation and speech recognition where both the input and output
sequence length is not fixed. 
A traditional architecture of seq2seq predictive networks has 
{\it two} RNs and a predictor \cite{Sutskever2014}. The first RN maps the whole input sequence 
of the length $m$
into a single inner state vector $\bm s_m$, this vector is repeatedly ($k$ times) fed into the second RN
and {\it each} its output $\bm \sigma_i$ is used by the predictor to generate
the output sequence. In this approach the same output $\bm \sigma_i$ should be also retained as current inner 
state of the second RN to be updated at the next input of the vector $\bm s_m$. 
This means that one has to maintain several copies of the vector $\bm s_m$ 
as well as to reserve memory for the inner states $\bm \sigma_i$ of the second RN. Again it is 
not clear whether these conditions can be satisfied in the neuroscience context.

In this manuscript the author  first considers the traditional seq2seq
algorithm with two RNs and a predictor. It is shown that if the predictive 
network employing such an algorithm is well trained (i.e., the
deviation of the predicted value sequence from the
ground truth one is negligibly small) there exists a nontrivial 
functional equation relating the parameters of both  RNs and the predictor.
In other words, knowledge of the parameters of the first RN and the predictor
determines the parameters of the second RN. This relation can be used to 
improve the prediction quality of the whole network.

The author also shows that there exists a natural extension of the ML approach
reported in \cite{Rub2020a} that allows design of a seq2seq ML algorithm. The numerical simulations
show that this algorithm is robust and its predictive quality is not worse and in some cases
is even better than demonstrated by the traditional one. The same time it has 
a clear advantage from the point of view of its application in the natural neural systems.

\section{Traditional seq2seq RNN}
\label{tradition}
The traditional seq2seq recurrent network architecture is actually 
comprised of {\it two} independent RNs and the linear predictor.
The input sequence $\bm X = \{\bm x_i\},\ 1 \le i \le m,$ of $d$-dimensional elements $\bm x_i$ is fed into
the first  RN  made of $n_1$ neurons that generates the corresponding 
states sequence $\bm S = \{\bm s_i\},\ 1 \le i \le m$. The elements of $\bm S$ are $n_1$-dimensional
vectors $\bm s_{i}$ representing inner states
of RN computed using a recurrent relation
\be
\bm s_{i} = \bm F_1(\bm x_i, \bm s_{i-1}),\quad
\bm s_0 = \bm 0,
\label{F1}
\ee
which describes a simple rule -- the current inner state $\bm s_{i}$ of the RN depends on the 
previous inner state $\bm s_{i-1}$ and the current input signal $\bm x_{i}$. This rule corresponds to an
assumption that the neural network does not store its state but just updates it
with respect to the submitted input signal and its previous state.
The final state $\bm s_{m}$ is replicated $k$ times producing the 
input sequence $\bm Y = \{\bm y_i\},\ \bm y_i = \bm s_m,\ 1 \le i \le k$ that is fed into the
second RN which $n_2$-dimensional inner states $\bm \sigma_i$  are determined by the relation 
\be
\bm \sigma_{i} = \bm F_2(\bm y_i, \bm \sigma_{i-1}) =\bm F_2(\bm s_m, \bm \sigma_{i-1}) ,\quad
\bm \sigma_0 = \bm 0.
\label{F2}
\ee
{\it All} inner states $\bm \sigma_i$ are linearly transformed by the 
predictor P to produce
\be
\bar {\bm x}_{m+i} = \bm P(\bm \sigma_{i}),
\quad 1 \le i \le k,
\label{P}
\ee
a sequence of $k$ predicted $d$-dimensional values $\bar {\bm x}_{m+i}$
approximating the ground truth ones  $\bar {\bm x}_{m+i} \approx {\bm x}_{m+i}$.
We assume that the predictive network is well trained, i.e., 
the deviations between $\bar {\bm x}_{m+i}$ and ${\bm x}_{m+i}$ can be neglected.
This $m$-to-$k$ network is a generalization of $m$-to-$1$ predictive networks 
that employs only a single recurrent network $F_1$ and the predictor P.
The described algorithm requires memory sufficient to hold $k$ states 
$\bm \sigma_i$  in proper order to be transformed into the predicted sequence of $\bar {\bm x}_{m+i}$.


\section{Dependence of the recurrent networks}
\label{dependence}
Consider first few prediction rounds of the expanding window algorithm. 
In what follows the round number $j$ 
is denoted as the superscript of the corresponding quantity.

\underline{Round $1$.}
The input sequence $\bm X^1 = \{\bm x_i\},\ 1 \le i \le m$. 
The first RN state sequence $\bm S^1 = \{\bm s_i\},\ 1 \le i \le m$
produced by $\bm s_{i} = \bm F_1(\bm x_i, \bm s_{i-1})$. 
The second RN inner states are computed by
$\bm \sigma_{i}^1 = \bm F_2(\bm s_m, \bm \sigma_{i-1}^1)$ and used further to generate
\be
\bar {\bm x}_{m+i}^1 = \bm P(\bm \sigma_{i}^1),
\quad 1 \le i \le k,
\label{P1}
\ee

\underline{Round $2$.}
The input sequence $\bm X^2$ is produced by appending the first predicted element 
$\bar {\bm x}_{m+1}^1 \approx \bm x_{m+1}$
to the sequence $\bm X^1$. Assuming that the added element $\bar {\bm x}_{m+1}^1$
in $\bm X^2$ can be replaced by the 
ground truth value $\bm x_{m+1}$ we have 
 $\bm X^2 = \{\bm x_i\},\ 1 \le i \le m+1$.
The last element $\bm s_{m+1}$ of the first RN state sequence $\bm S^2 = \{\bm s_i\},\ 1 \le i \le m+1$
 is replicated and used as input to the second RN
$\bm \sigma_{i}^2 = \bm F_2(\bm s_{m+1}, \bm \sigma_{i-1}^2)$ and used further to generate
\be
\bar {\bm x}_{m+1+i}^2 = \bm P(\bm \sigma_{i}^2),
\quad 1 \le i \le k,
\label{P2}
\ee

\underline{Round $3$.}
The input sequence $\bm X^3$ is produced by appending the second predicted element 
$\bar {\bm x}_{m+2}^2 \approx \bm x_{m+2}$
to the sequence $\bm X^2$ and  we have 
 $\bm X^3 = \{\bm x_i\},\ 1 \le i \le m+2$.
The last element $\bm s_{m+2}$ of the first RN state sequence $\bm S^3 = \{\bm s_i\},\ 1 \le i \le m+2$
 is replicated and used as input to the second RN
$\bm \sigma_{i}^3 = \bm F_2(\bm s_{m+2}, \bm \sigma_{i-1}^3)$ and used further to generate
\be
\bar {\bm x}_{m+2+i}^2 = \bm P(\bm \sigma_{i}^3),
\quad 1 \le i \le k,
\label{P3}
\ee

\underline{Round $k$.}
The input sequence $\bm X^k$ is produced by appending the $(k-1)$-th predicted element 
$\bar {\bm x}_{m+k-1}^2 \approx \bm x_{m+k-1}$
to the sequence $\bm X^{k-1}$ and  we have 
 $\bm X^k = \{\bm x_i\},\ 1 \le i \le m+k-1$.
The last element $\bm s_{m+k-1}$ of the first RN state sequence $\bm S^k = \{\bm s_i\},\ 1 \le i \le m+k-1$
 is replicated and used as input to the second RN
$\bm \sigma_{i}^k = \bm F_2(\bm s_{m+k-1}, \bm \sigma_{i-1}^k)$ and used further to generate
\be
\bar {\bm x}_{m+k-1+i}^k = \bm P(\bm \sigma_{i}^k).
\quad 1 \le i \le k,
\label{Pk}
\ee

From (\ref{P1}) and (\ref{P2}) it follows that the element $\bar {\bm x}_{m+2}$ 
predicted in both the first ($j=1$) and the second ($j=2$) prediction rounds.
Compare the values $\bar {\bm x}_{m+2}^j$ for $j=1,2$.
From (\ref{P1}) we obtain $\bar {\bm x}_{m+2}^1 = \bm P(\bm \sigma_2^1),$
where $\bm \sigma_2^1 = \bm F_2(\bm s_m, \bm \sigma_1^1)$,
and $\bm \sigma_1^1 = \bm F_2(\bm s_m, \bm 0)$, so that
\be
\bar {\bm x}_{m+2}^1 = \bm P(\bm F_2(\bm s_m, \bm F_2(\bm s_m, \bm 0))).
\label{x_{m+2}^1}
\ee
On the other hand (\ref{P2}) leads to 
 $\bar {\bm x}_{m+2}^2 = \bm P(\bm \sigma_1^2),$
where $\bm \sigma_1^2 = \bm F_2(\bm s_{m+1}, \bm 0)$,
and we obtain 
\be
\bar {\bm x}_{m+2}^2 = \bm P(\bm F_2(\bm s_{m+1}, \bm 0)).
\label{x_{m+2}^2}
\ee
Using
$$
\bm s_{m+1} = \bm F_1(\bar {\bm x}_{m+1}^1,\bm s_m) = 
\bm F_1(\bm P(\bm \sigma_{m+1}^1),\bm s_m) = 
\bm F_1(\bm P(\bm F_2(s_{m},\bm 0)),\bm s_m).
$$
in the above relation we arrive at
\be
\bar {\bm x}_{m+2}^2 = \bm P(\bm F_2(\bm F_1(\bm P(\bm F_2(s_{m},\bm 0)),\bm s_m), \bm 0)).
\label{x_{m+2}^2a}
\ee
For the well trained predictive network the 
values $\bar {\bm x}_{m+2}^j$ with $j=1$ and $j=2$
should be very close to each other and we 
assume them to be equal. As the predictor P performs 
in both cases the same linear transformation we conclude that
$\bm \sigma_2^1 = \bm \sigma_1^2$ and we arrive at
\be
\bm F_2(\bm s_{m}, \bm F_2(\bm s_{m}, \bm 0)) =
\bm F_2(\bm F_1(\bm P(\bm F_2(\bm s_{m},\bm 0)),\bm s_{m}), \bm 0).
\label{cond2}
\ee
Repeating the same steps for a pair of $\bar {\bm x}_{m+3}^j$ for $j=2$ and $j=3$
we find similar to (\ref{cond2})
\be
\bm F_2(\bm s_{m+1}, \bm F_2(\bm s_{m+1}, \bm 0)) =
\bm F_2(\bm F_1(\bm P(\bm F_2(\bm s_{m+1},\bm 0)),\bm s_{m+1}), \bm 0).
\label{cond3}
\ee
By induction the following relation holds 
$$
\bm F_2(\bm s_{i}, \bm F_2(\bm s_{i}, \bm 0)) =
\bm F_2(\bm F_1(\bm P(\bm F_2(\bm s_{i},\bm 0)),\bm s_{i}), \bm 0),
\quad 
m \le i \le m+k-1.
$$
As the input sequence generating the inner values $\bm s_i$ can be selected from
a large number of samples we
conclude that the above relation must also be valid for {\it every} hidden 
vector $\bm s$ corresponding to any input value $\bm x$ that belongs to sequences used for 
network training
\be
\bm F_2(\bm s, \bm F_2(\bm s, \bm 0)) =
\bm F_2(\bm F_1(\bm P(\bm F_2(\bm s,\bm 0)),\bm s), \bm 0).
\label{condgen}
\ee
This implies that for the well trained seq2seq predictive network
there exists a set of nontrivial relations (\ref{condgen}). Given the 
function $\bm F_1$ determining the first RN and the linear transformation 
$\bm P$ for the predictor the relations (\ref{condgen}) restrict and
actually {\it define} the  function $\bm F_2$. In other words, the 
RNs are {\it not independent} --
the functional equation (\ref{condgen}) represents 
a condition on parameters of the ideal predictive network
and can be viewed as a tool for network 
improvement. It can be done as follows -- first the 
network is trained using standard backpropagation algorithm
fixing the parameters of all three
components of the network. Then the parameters
of any two of the three components (preferentially, the predictor and the first RN generating $\bm s$ values)
are fixed and the parameters of the remaining RN are tuned
to satisfy the relation (\ref{condgen}) as good as possible.

\section{Memoryless algorithm}
\label{ML}
The traditional seq2seq network architecture and the corresponding algorithm lead 
to a specific memory requirements that can be easily implemented {\it in silico}
but in the author opinion is quite difficult to satisfy in natural neural networks.

First, one has to produce $k$ exact copies of $\bm s_m$ and feed them one by one into the 
second RN. It can be done if existence time of the inner state $\bm s_m$ is equal or larger
than an interval required to process copies of this state $k$ times through the second RN.
Second, each inner state $\bm \sigma_i$ of the second RN should be used as an input in two
independent processes -- nonlinear transformation (\ref{F2}) and
linear transformation (\ref{P}) of the predictor P. It can be done by making a copy of  $\bm \sigma_i$ 
before feeding it into the predictor.

On the other hand it is possible to simplify network architecture and use a memoryless (ML) algorithm 
introduced recently \cite{Rub2020a,Rub2020b} for the $m$-to-$1$ predictive networks.
The essence of the method is that for the well trained RN (with $\bar{\bm x}_{m+1} \approx \bm x_{m+1}$)
one can produce a sequence of $\bm s_{m+i+1}, \ 0\le i \le p-1$ using 
a simple relation for the nonlinear transformation $\bm F$ of the single RN:
\be
\bm s_{m+i+1} = \bm F(\bm P(\bm s_{m+i}),\bm s_{m+i}) = 
\bm F(\bar{\bm x}_{m+i+1},\bm s_{m+i}) ,
\label{MLorig}
\ee
 without constructing new input sequences $\bm X^{i+1}$
required by the EW or MW approach.
Notice that in ML algorithm a computation of each new predicted element 
$\bar{\bm x}_{m+i+1}=\bm P(\bm s_{m+i})$ naturally leads to $\bm s_{m+i+1}$ used for prediction 
of the next element $\bar{\bm x}_{m+i+2}$ while {\it no memory} is required in
this recursive process.

The relation (\ref{MLorig}) allows to produce 
sequence of $k$ predicted values $\bar{\bm x}_{m+i}, \ 1\le i \le k$,
compare it to the sequence of the ground truth
values ${\bm x}_{m+i}, \ 1\le i \le k$ and compute the training error $E_1$ (defined below)
 used in backpropagation training algorithm.

After the network is trained to predict $k$ values it is easy to 
extend it for prediction of a sequence having $pk$ of 
elements reusing  (\ref{MLorig}) recursively, and one can define a prediction error $E_p$
defined as 
\be
E_p^2 = \frac{1}{kp}\sum_{i=1}^{kp} \| \bar{\bm x}_{m+i}-{\bm x}_{m+i} \|^2\;,
\label{error}
\ee
where $\| \bm v \|$ denotes an Euclidean norm ($L_2$-norm) of vector $\bm v$.
The training error $E_1$ is a particular case of (\ref{error}) for $p=1$.

\section{Numerical simulations}
\label{numerics}
It is instructive to compare the two architectures of the 
seq2seq predictive networks described in the previous Sections.
First we consider the traditional algorithm (Section \ref{tradition}) and then
turn to the ML approach (Section \ref{ML}).

\subsection{Traditional seq2seq network}
\label{numerics_Trad}
As the traditional networks employ two RNs with the number of neurons
equal to $n_i,\ i=1,2,$ it is interesting to learn what 
ratio $r = n_1/n_2$ for fixed total number $n = n_1+n_2$
leads to the smallest error $E_p$ defined by (\ref{error}).
To address this problem we train networks
to predict the time series of the 
phase modulated 1D noisy signals -- sine wave $G_s(t) =  a \xi(t)  + A_0 + A \sin(2\pi t/T)$ and trapezoid wave
$$
G_t(t) = a \xi(t) + A_0 +
\left \{ 
\begin{array}{ll}
A t/r, & 0\le t < r, \\
A, & r \le t < r +w, \\
A(r + w + f - t)/f, & r+w \le t < r+w+f, \\
0, & r+w+f \le t < T=r+w+f+s,
\end{array}
\right.
$$
where $T$ is the wave period,
$a$ is the amplitude of white noise $\xi(t)$, $A_0$ is the offset and
$A$ is the wave amplitude. The phase modulation 
is implemented by following argument replacement $t \to t + \Delta \sin(2\pi t/s)$, where 
$\Delta$ is the amplitude of the phase modulation and $s$ defines
its periodicity.

The training set construction is performed as follows: 
for given function $G_s$ or $G_t$ we create a
set of points $G(t_i)$ with $t_i = i\times \delta t$ where $1 \le i \le 20000$  and
$\delta t = 0.01$ and noise amplitude $a=0.15$. The parameters of phase modulation
read $\Delta = 2,\ s=10$, while the trapezoid parameters 
are $r=f=0.1, \ w,s = 0.4$, so that $T=1$.
Then from each set a pairs of input $\bm I$ and output $\bm O$ sequences
are generated -- $\bm I$ contains the values $G(t_i)$ with $p \le i \le p+m-1$ and the 
ground truth sequence
$\bm O$ contains the values $G(t_i)$ with $p+m \le i \le p+m+k-1$.
We use $20 \le m \le 80$, while the length $k$ of the output sequence 
is equal to $k=10$. For each type of signals 4000 training samples are produced
and merged into a single training set. The networks are trained 
using the Adam algorithm for $50$ epochs 
with $20\%$ of data used as a validation set.

\begin{figure}[h!]
\begin{center}
\begin{tabular}{cc}
\psfig{figure=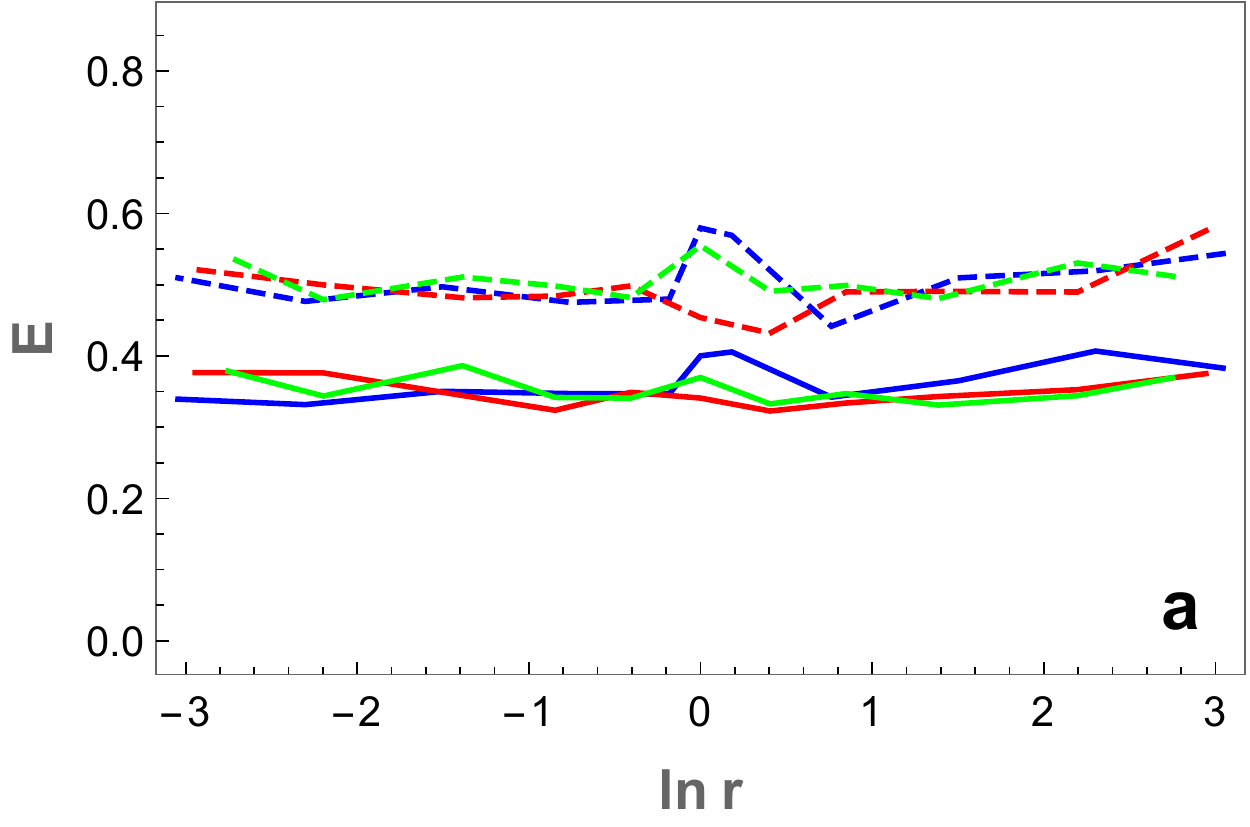,width=7.5cm} & 
\psfig{figure=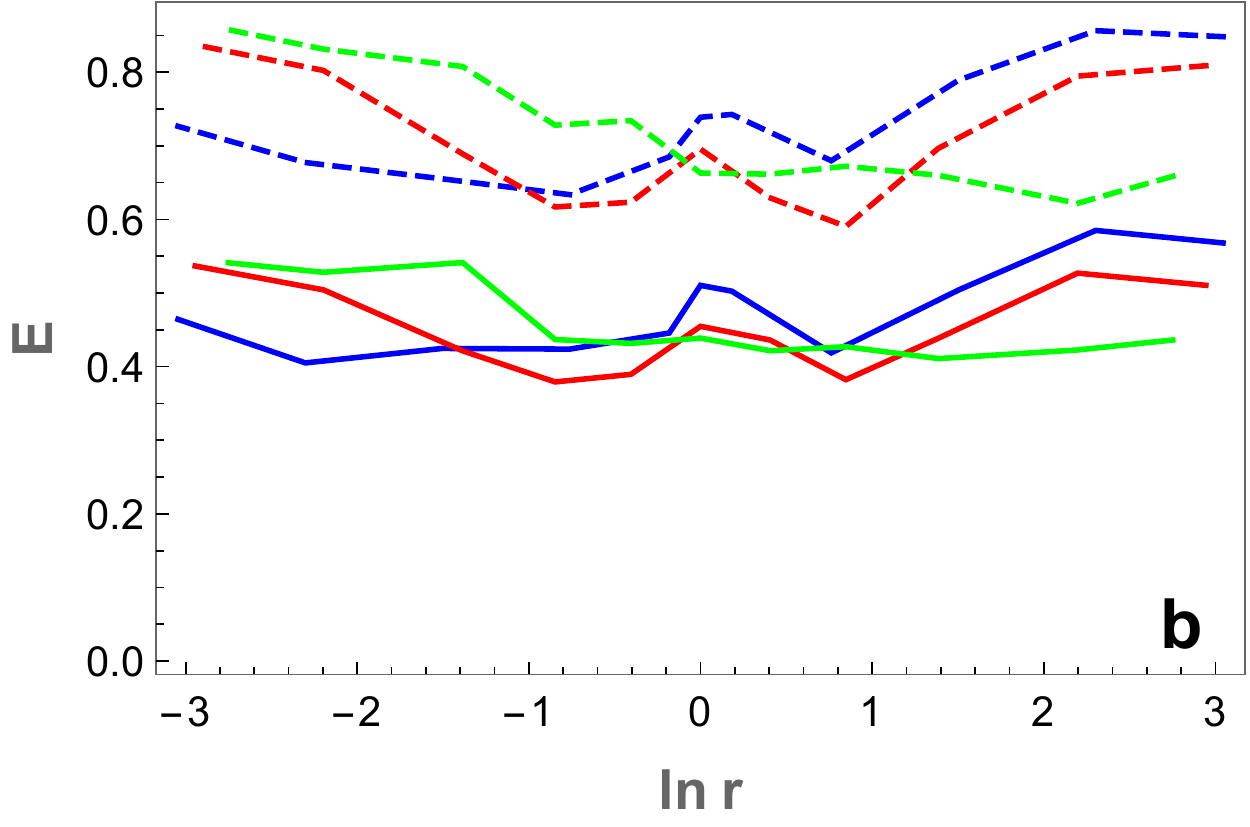,width=7.5cm} 
\end{tabular}
\caption{
Dependence of the error $E$ on logarithm $\ln r$ of the ratio $r=n_1/n_2$ for 
(a) sine and (b) trapezoid phase modulated wave with added noise
of amplitude $a=0.15$.
The total number of neurons $n=n_1+n_2$ is $n=50$ (green), $n=100$ (blue) and $n=220$ (red).
The length of the input sequence $m=70$ and the predicted sequence size is $k=10$.
The error values are found as an average of $1000$ randomly selected input sequences.
Both RNs were selected to be the basic (vanilla) recurrent networks.
The solid and dashed curves represent $kp=10$ and $kp=40$ total number of predicted points
respectively.
}
\label{trad1}
\end{center}
\end{figure}

The analysis of the simulation results are presented in Fig. \ref{trad1}. 
First we observe that the sine wave prediction quality (Fig. \ref{trad1}a) does not
depend significantly on the total number $n$ of neurons. On the 
other hand for trapezoid wave (Fig. \ref{trad1}b) both the ratio $r$ and the total neuron number $n$
influence the training and prediction errors.
We observe in this case that 
when the total number $n$ of neurons is small the prediction quality improves for 
larger ratios $r$ (solid curves). These trends
are reproduced when one recursively repeats the prediction algorithm (dashed curves).
When the total number of neurons is large ($n=220$) the error demonstrates average growth for increasing $r$ with 
local minima and maximum around $r \sim 1$. Finally in the intermediate case $n=100$ the minimal error is
observed for the ratios $r \approx 1$.

\begin{figure}[h!]
\begin{center}
\begin{tabular}{cc}
\psfig{figure=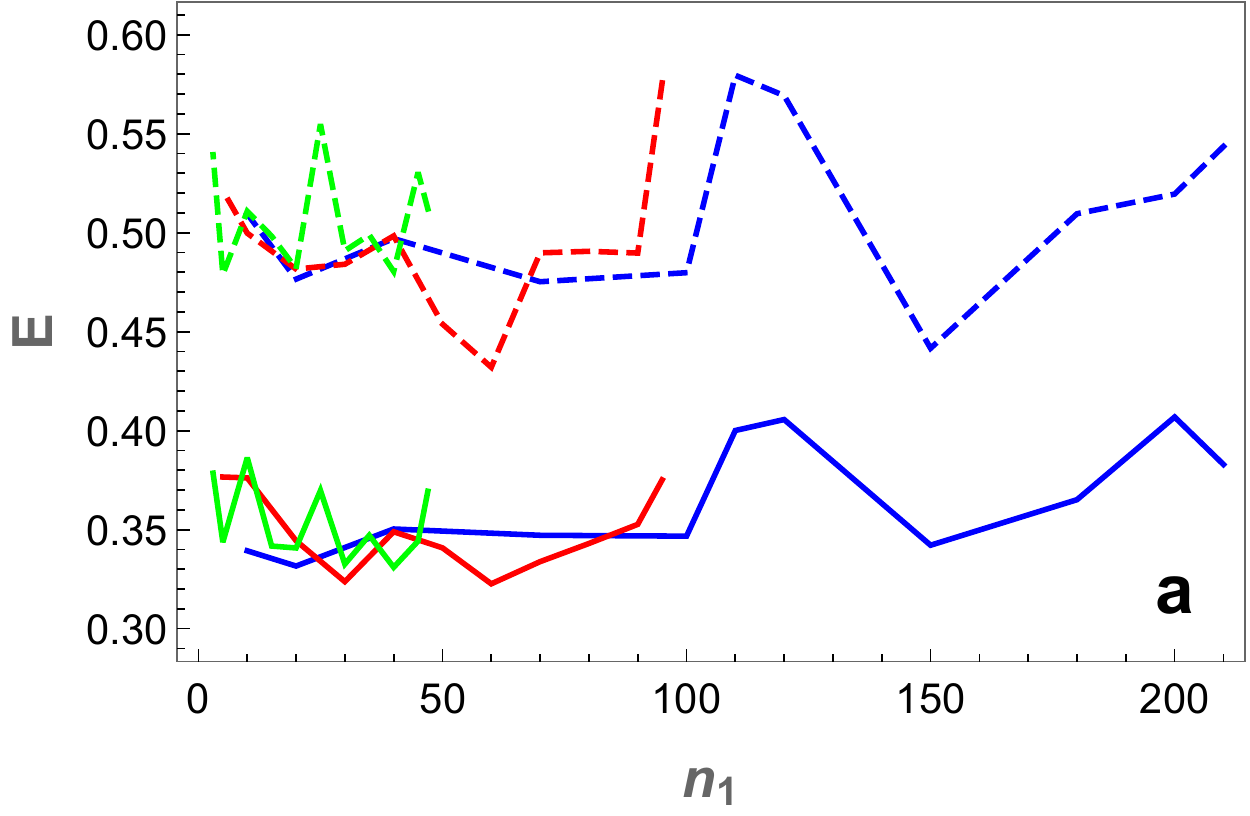,width=7.5cm} & 
\psfig{figure=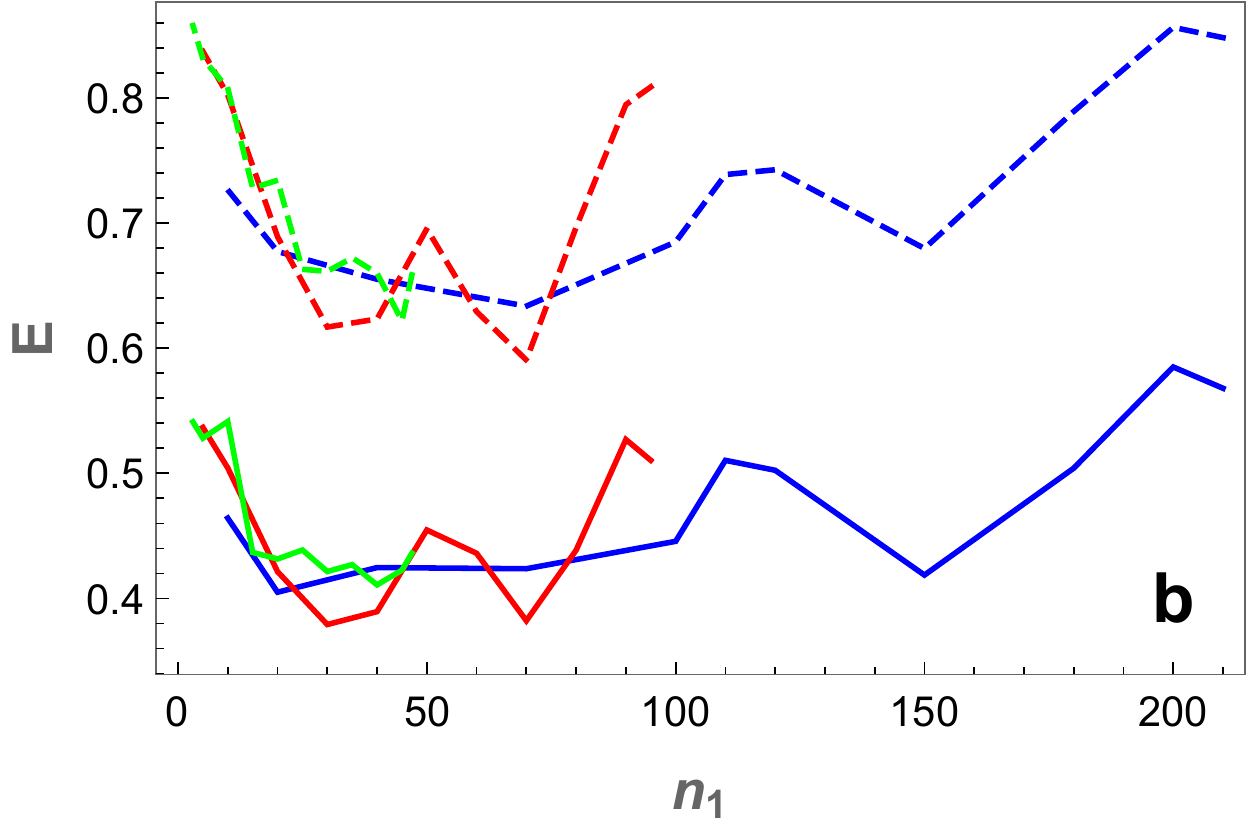,width=7.5cm} 
\end{tabular}
\caption{
Dependence of the error $E$ on the number $n_1$ of neurons in the first RN for
(a) sine and (b) trapezoid noisy wave.
The total number of neurons $n=n_1+n_2$ is $n=50$ (green), $n=100$ (blue) and $n=220$ (red).
The solid and dashed curves represent $kp=10$ and $kp=40$ total number of predicted points
respectively. All other parameters are as in Fig. \ref{trad1}.
}
\label{trad1n}
\end{center}
\end{figure}

Another important trend (Fig. \ref{trad1n}) demonstrates that the error 
$E$ dependence on the number $n_1$ of the neurons in the first basic RN 
is on average the same (with some local deviations) for different total number $n$ of the neurons in predictive network.
We observe that for the sine wave the error does not change significantly for $n \le 50$ and starts to increase with 
$n \ge 100$. In case of trapezoid wave the error decreases when $n$ is below $30$ but for larger $n$ it starts to increase
but this behavior is nonmonotonic.

\subsection{Memoryless seq2seq network}
\label{numerics_ML}
\begin{figure}[h!]
\begin{center}
\begin{tabular}{cc}
\psfig{figure=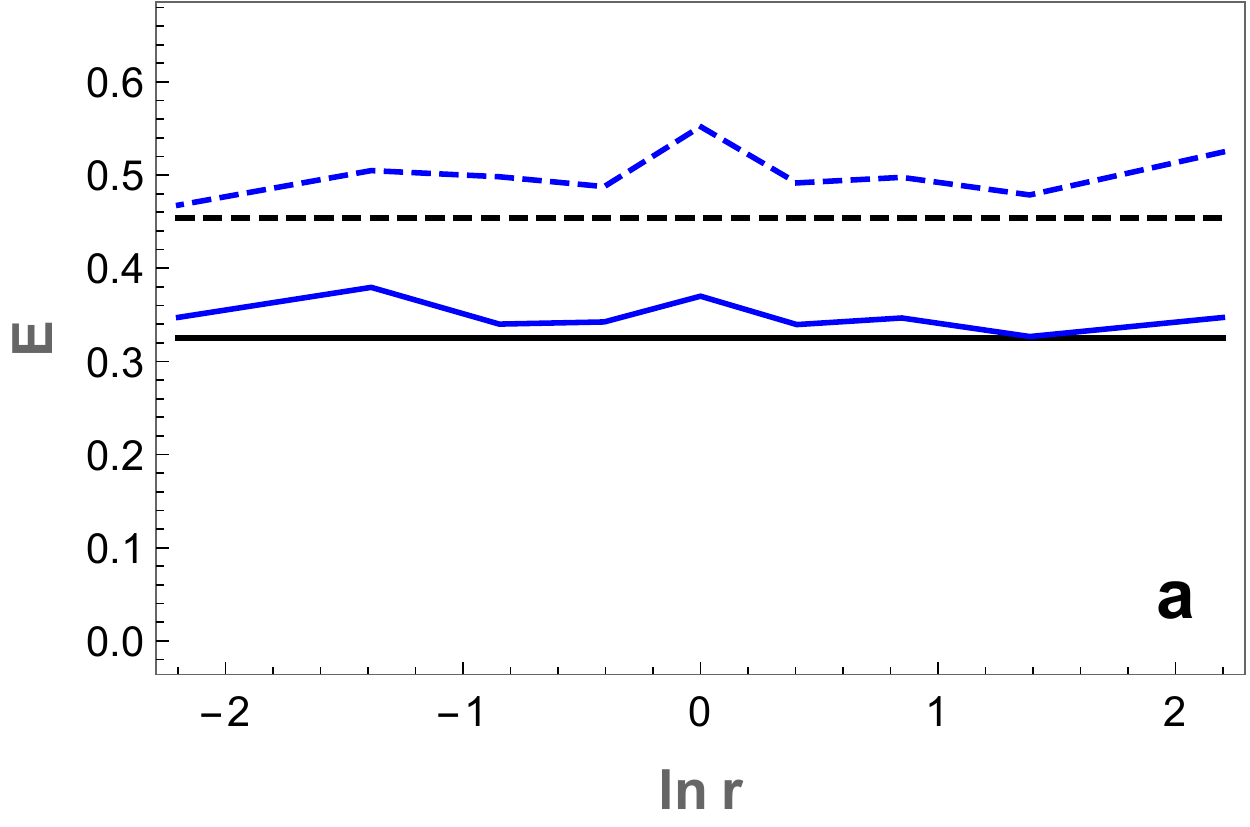,width=7.5cm} &
\psfig{figure=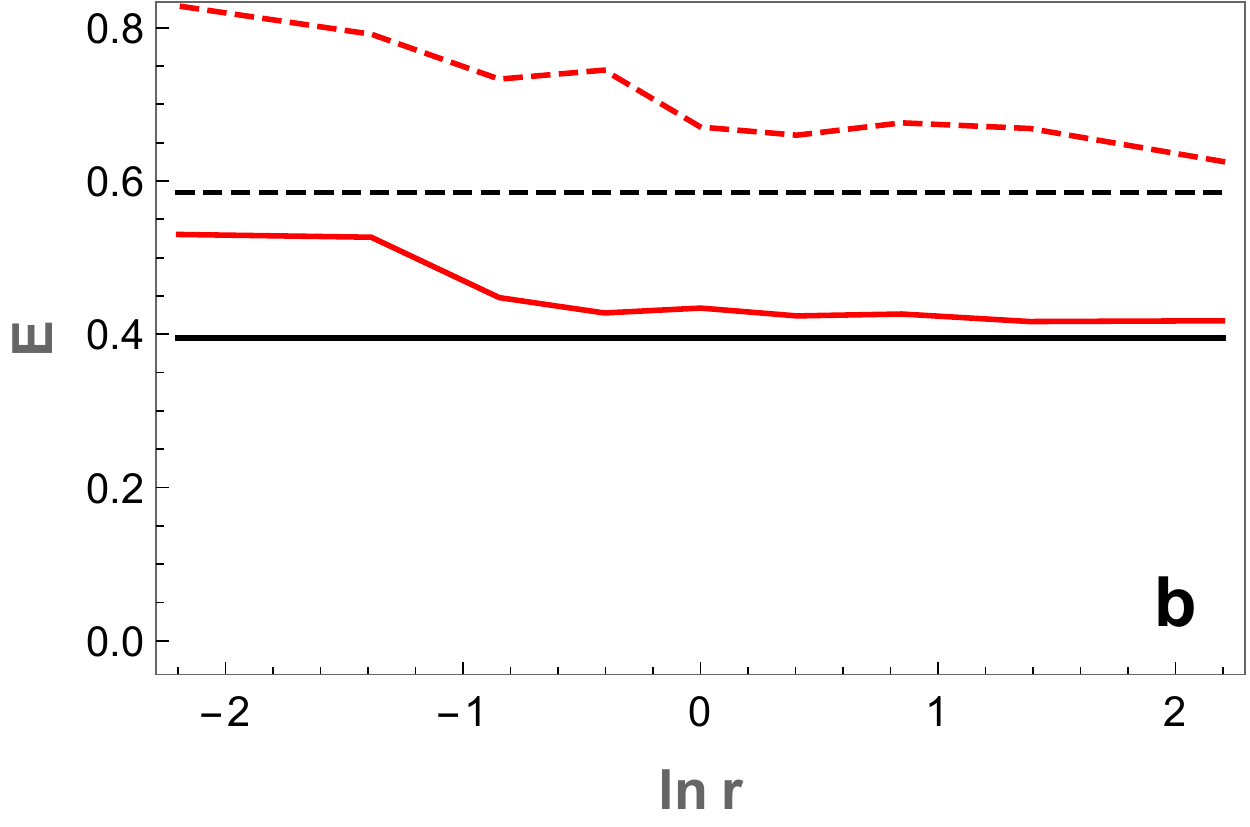,width=7.5cm} 
\end{tabular}
\caption{
Dependence of the error $E$ on logarithm $\ln r$ of the ratio $r=n_1/n_2$ for 
the total number of neurons $n=50$
compared to the error value in ML networks with the same $n$.
Comparison for (a) sine and (b) trapezoid  phase modulated wave with added noise
of amplitude $a=0.15$.
The length of the input sequence $m=70$ and the predicted sequence size is $k=10$.
The error values are found as an average of $1000$ randomly selected input sequences.
Blue (a) and red (b0 curves correspond to $G_s$ and $G_t$ respectively; the black curve
describes the ML network error.
The solid and dashed curves represent $kp=10$ and $kp=40$ total number of predicted points
respectively.
}
\label{trad2}
\end{center}
\end{figure}

To compare the prediction quality of the traditional and
the memoryless networks we construct a predictive
network with a single basic RN having $n=50$ neurons and train it 
on the same data set that was used for the traditional one.
We observe that the error estimates in ML networks
are consistently lower than those for the traditional one (Fig. \ref{trad2}).
The same time the trends for the sine and trapezoidal noisy waves are opposite --
for the sine wave the ML algorithm reports smaller error for medium and large ratios (Fig. \ref{trad2}a), 
while  for the trapezoidal signal it becomes significantly lower at small ratios (Fig. \ref{trad2}b).

\begin{figure}[h!]
\begin{center}
\begin{tabular}{cc}
\psfig{figure=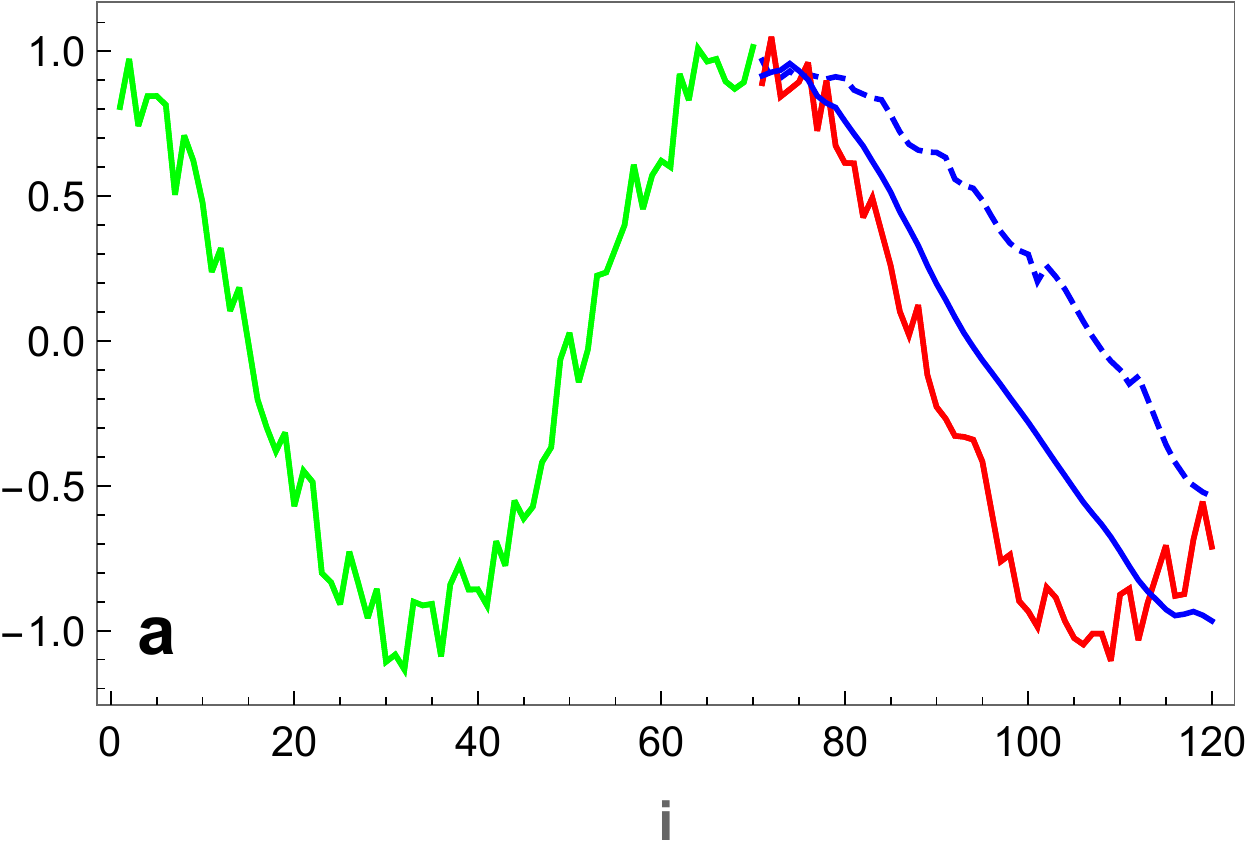,width=7.5cm} &
\psfig{figure=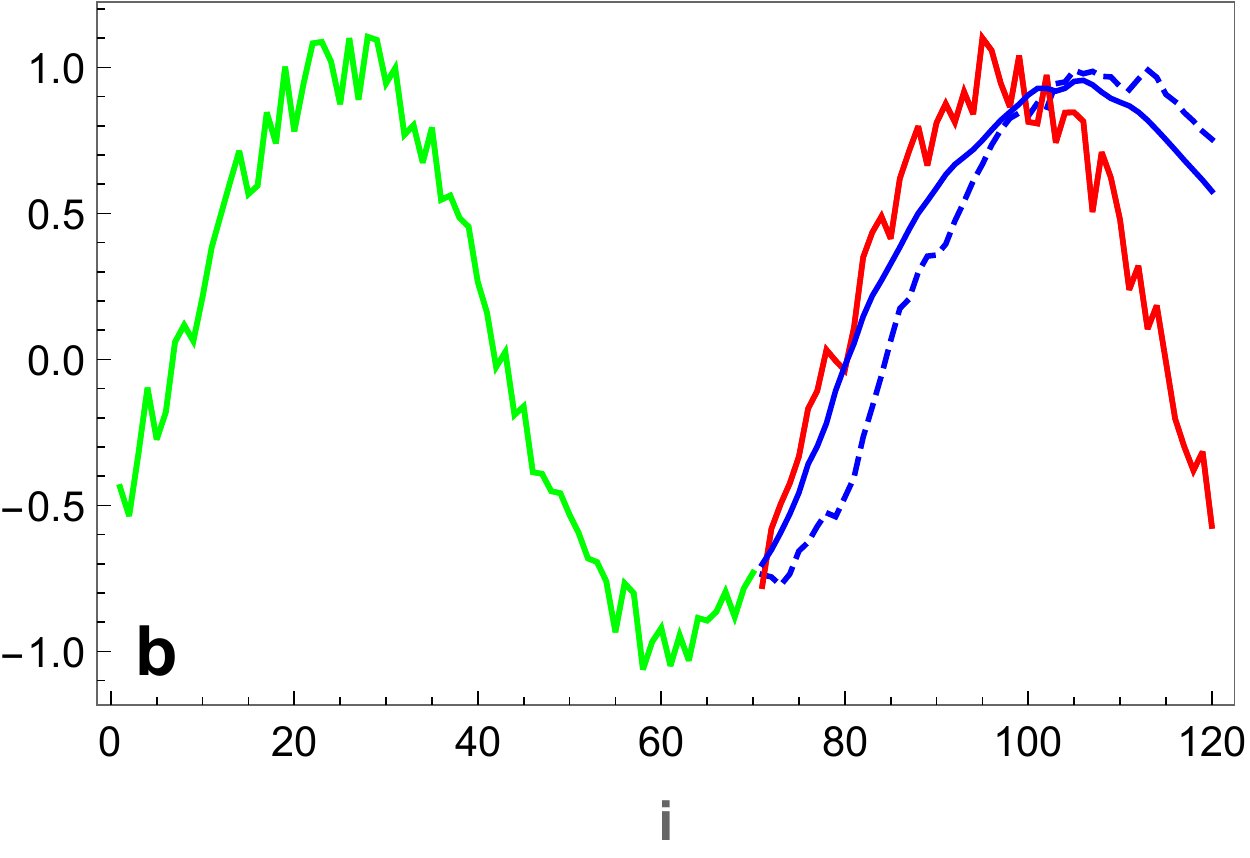,width=7.5cm} \\  
\psfig{figure=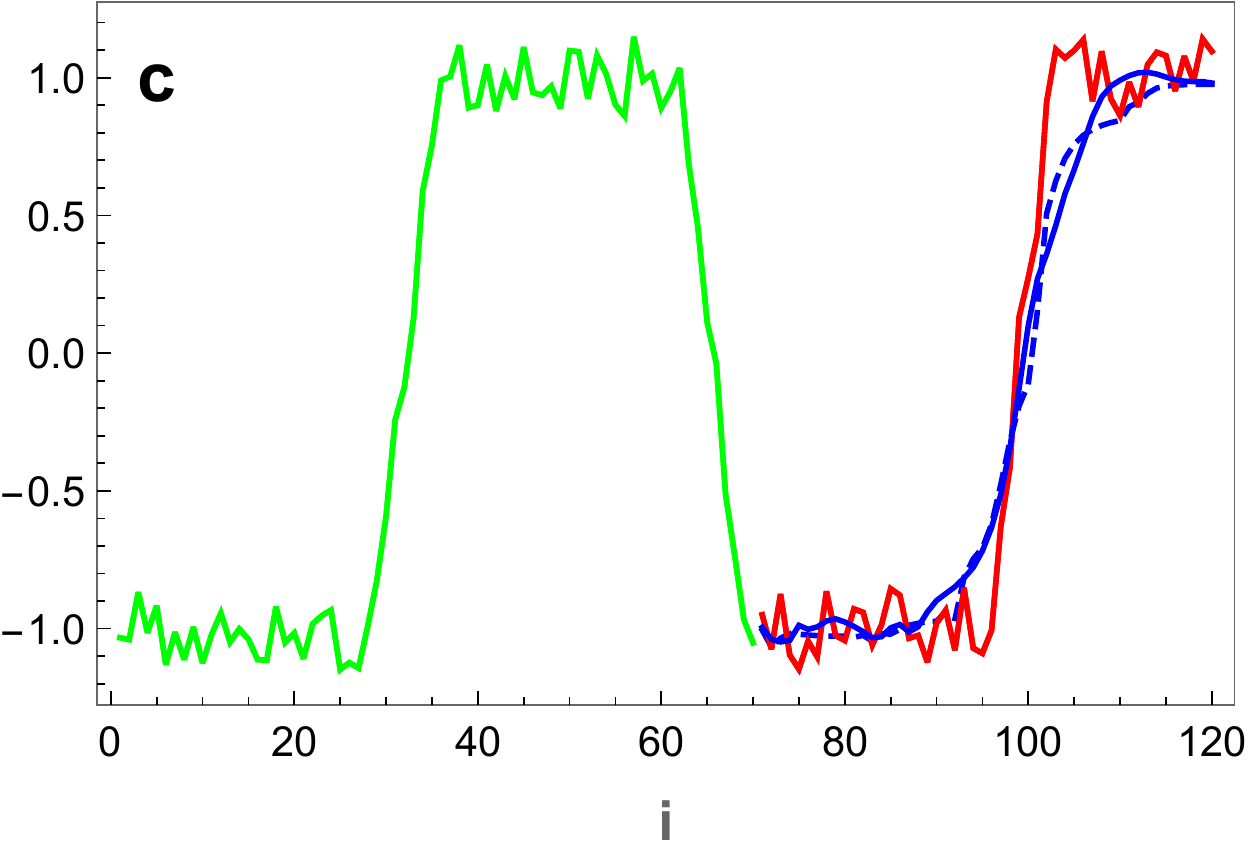,width=7.5cm} &
\psfig{figure=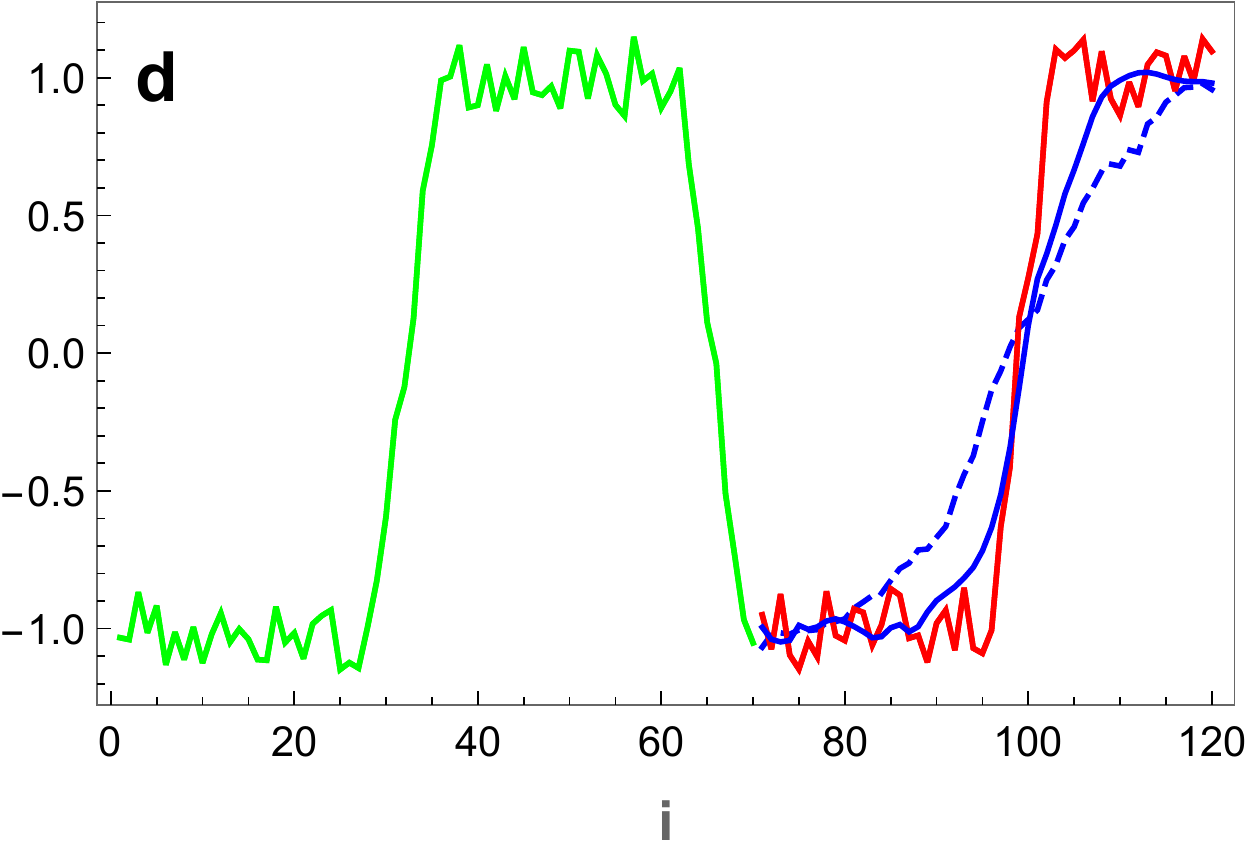,width=7.5cm} 
\end{tabular}
\caption{
Comparison of the ground truth continuation (red) of the input noisy phase modulated sine (a,b) and
trapezoid (c,d) wave sequence (green)
to the predictions computed by ML (solid blue) and traditional (dashed blue) algorithms
in the network with the total number of neurons $n=50$.
The length of the input sequence $m=70$ and the predicted sequence size is $kp=40$.
The ratio $r$ of the traditional network is $r=4$ (a,c) and $r=1/4$ (b,d).
}
\label{compareSin}
\end{center}
\end{figure}

We illustrate these observations in Fig. \ref{compareSin} 
showing the input sequence curve, its 
ground truth continuation and the predicted curve obtained by employing 
both algorithms in the networks with $n=50$.
We confirm that for large values of $r$ the ML network predicts the sine wave better than the
traditional one.
On the other hand  the ML network predicts much better the trapezoid wave better than the
traditional one for smaller ratios while for large ratios the predicted curves effectively coincide.

\section{Discussion}
\label{discussion}
In this manuscript the author considers the traditional architecture and 
training algorithm of seq2seq predictive network that includes two 
RNs and a predictor. It appears that for this network the parameters
of the second RN depend on those defining the first RN and the predictor.
This dependence has a form of a functional vector equation satisfied 
for a very large number of the vector arguments $\bm s_m$.  These 
vectors depend both of the parameters of the first RN and the sample input sequence, i.e., 
on the time series to be predicted. 

It is important to underline that the 
established functional equation corresponds to the {\it ideally} trained 
predictive network and cannot be satisfied for all arguments.
The same time it can serve as a tool to improve the predictive power of
the network in the following manner. First the traditional network
is trained using standard algorithms. Then
for the fixed parameters of the first RN $\bm F_1$ and the predictor $\bm P$
one performs tuning of parameters of the second RN $\bm F_2$ using arguments
$\bm s_m$ generated by feeding the input sequences from the training set
into the first RN. The choice of the tuning algorithm
will be discussed elsewhere.

The traditional seq2seq algorithm requiring memory
to preserve the replicated inner state $\bm s_m$ might be difficult 
to implement in neuroscience context. To overcome this difficulty one can 
use an alternative memoryless (ML) algorithm being an extension of the 
algorithm proposed recently in \cite{Rub2020a,Rub2020b}. The network 
implementing this approach employs only a single RN and a predictor is
shown to successfully predict the phase modulated noisy periodic signals.
The comparison to the traditional seq2seq networks demonstrates that 
the ML network has lower error, i.e., higher prediction quality.

\section*{Acknowledgements}
The author wishes to thank Jay Unruh for fruitful discussions.


\end{document}